\documentclass[runningheads]{llncs}
\usepackage{graphicx}
\usepackage{custom_commands}
\usepackage{custom_aliases}
\usepackage{comment}
\usepackage{todonotes}
\usepackage{graphicx, amsmath, amssymb}
\usepackage{caption}
\usepackage{subcaption}
\usepackage{bm}

\usepackage[normalem]{ulem}
\begin{document}
\title{Efficient Approximation of\\
Expected Hypervolume Improvement\\using Gauss-Hermite Quadrature}
%\\
%Efficiency of Gauss Hermite Approximation for Estimating Expected Hypervolume Improvement \\
% Gauss Hermite Approximation for Estimating Expected Hypervolume Improvement}
%
\titlerunning{Expected Hypervolume Improvement using Gauss-Hermite Quadrature}
\author{Alma Rahat\inst{1}\orcidID{0000-0002-5023-1371} \and
Tinkle Chugh\inst{2}\orcidID{0000-0001-5123-8148} \and
Jonathan Fieldsend\inst{2}\orcidID{0000-0002-0683-2583}
\and Richard Allmendinger\inst{3}\orcidID{0000-0003-1236-3143}
\and
Kaisa Miettinen\inst{4}\orcidID{0000-0003-1013-4689}}
\authorrunning{Rahat et al.}
\institute{Swansea University, Swansea, SA1 8EN, U.K. \\
\email{a.a.m.rahat@swansea.ac.uk}\and
University of Exeter, Exeter EX4 4QD, U.K.
\\ \email{\{T.Chugh,J.E.Fieldsend\}@exeter.ac.uk} \and
The University of Manchester, Manchester M15 6PB, U.K.\\\email{richard.allmendinger@manchester.ac.uk}
\and
University of Jyvaskyla, Faculty of Information Technology\\ P.O. Box 35 (Agora), FI-40014 University of Jyvaskyla, Finland\\ \email{kaisa.miettinen@jyu.fi}
}
\maketitle              % typeset the header of the contribution
\begin{abstract}
% Multi-objective Bayesian optimisation (BO) has been widely used to problems with computationally expensive objective functions. The expected hypervolume improvement (EHVI) is a popular and efficient acquisition function in multi-objective BO. However, a major drawback pf EHVI is that there is no closed form expression when a multi-task surrogate model is used to quantify the correlation between objective functions. An alternative is to use Monte Carlo sampling to approximate the EHVI, but this approach is still limited by the computational cost of the required simulations. This work proposes an alternative to estimating the EHVI, by employing a Gauss-Hermite (G-H) approach. It is computationally cheaper than Monte Carlo sampling and achieves a reasonable approximation accuracy. Our experimental study on benchmark problems shows the potential of the G-H approach. We compare the accuracy of the G-H approach to the closed form expression and Monte Carlo sampling, and further illustrate its use in multi-objective BO examples. 
Many methods for performing multi-objective optimisation of computationally expensive problems have been proposed recently. Typically, a probabilistic surrogate for each objective is constructed from an initial dataset. The surrogates can then be used to produce predictive densities in the objective space for any solution. Using the predictive densities, we can compute the expected hypervolume improvement (EHVI) due to a solution. Maximising the EHVI, we can locate the most promising solution that may be expensively evaluated next. There are closed-form expressions for computing the EHVI, integrating over the multivariate predictive densities. However, they require partitioning the objective space, which can be prohibitively expensive for more than three objectives. Furthermore, there are no closed-form expressions for a problem where the predictive densities are dependent, capturing the correlations between objectives. Monte Carlo  approximation is used instead in such cases, which is not cheap. Hence, the need to develop new accurate but cheaper approximation methods remains. Here we investigate an alternative approach toward approximating the EHVI using Gauss-Hermite quadrature. We show that it can be an accurate alternative to Monte Carlo for both independent and correlated predictive densities with statistically significant rank correlations for a range of popular test problems.

\end{abstract}

% \input{main_body}

% \vspace{-1em}
\section{Introduction}

Many real-world optimisation problems have multiple conflicting objectives \cite{deb2014multi,miettinen2012nonlinear,applications}. In many cases, these objective functions can take a substantial amount of time for one evaluation. For instance, problems involving computational fluid dynamic simulations can take minutes to days for evaluating a single design (or decision vector/candidate solution) \cite{allmendinger2012tuning,Chugh_Valtra}. Such problems do not have analytical or closed-form expressions for the objective functions %\sout{ (because these are impossible or too time-consuming to derive)} 
and are termed as black-box problems. To alleviate the computation time and  obtain solutions with few  expensive function evaluations, surrogate-assisted optimisation methods \cite{allmendinger2017surrogate,chughetal}, e.g.\ Bayesian optimisation (BO) \cite{Shahriari2016}, have been widely used. In such methods, a surrogate model (also known as a metamodel) is built on given data (which is either available or can be generated with some design of experiment technique~\cite{montgomery2017design}). %There are typically two ways to build model(s) in multi-objective optimisation. The first one is to scalarise the  multiple objectives into a single objective by using an appropriate scalarising function~\cite{mbore_george,Chugh_Scalarising_Functions} (e.g.\ weighted Tchebycheff in ParEGO~\cite{knowles}. %, or achievement scalarizing function in \cite{tabatabaeietal}). \mnote{TC: Not sure who wrote this, but its misleading. Both functions have their own purpose. They are not the same functions. Please, refer to Kaisa's book or simply talk to her.} 
%The second is to 
If one builds independent models for each objective function \cite{emmerich2011hypervolume,yang2015expected}, %. These methods do not directly consider 
the correlation between the objective functions is not directly considered. 
%when building a model. 
Multi-task surrogates~\cite{bonilla2007multi,shah2016pareto} have been used recently to consider the correlation. %between the objective functions.
%\tc{TC: I thin this needs further elaboration}

In BO, %\sout{a popular surrogate-assisted optimisation approach for expensive problem,}
the surrogate model is usually a Gaussian process (GP) because %One of the main benefits of using 
GPs %is that they 
provide uncertainty information in the approximation in addition to the point approximation. These models are then used in optimising an acquisition function (or infill criterion) to find the next best decision vector to evaluate expensively. The acquisition function usually balances the convergence and diversity. %\tc{TC: We need to be consistent, Use exploitation and exploration or convergence and diversity} 
Many acquisition functions have been proposed in the literature.  %. For instance,
%expected improvement \cite{Jones1998}, and expected hypervolume improvement \cite{Emmerich2005a}. %They have been used in multi-objective BO and its applications \cite{}. \tc{Alma's paper here} 
Here, we focus on using expected hypervolume improvement (EHVI)~\cite{emmerich2005single}, which has become a popular and well-studied acquisition function for expensive multi-objective optimisation largely due to its reliance on the hypervolume~\cite{zitzler2003performance} (the only strictly Pareto compliant indicator known so far). % computes by how much the collective hypervolume of a solution set improves due to adding a new solution to that set. This 
The EHVI relies on a predictive distribution of solutions (with either independent \cite{emmerich2005single} or correlated objective functions \cite{shah2016pareto}). An optimiser is used to maximise the EHVI to find a decision vector with maximum expected improvement in hypervolume. 
% hypervolume contribution. %\sout{by using distribution of solutions}
%\mnote{RA: Not sure what is meant by ``contribution by using distribution of solutions''}. 
The EHVI can be computed analytically for any number of objectives assuming the objective functions $f_1,\dots,f_m$ are drawn from independent GPs~\cite{emmerich2011hypervolume}. However, this computation is expensive for more than three objectives. Monte Carlo (MC) approximation of EHVI is often used instead in such cases but this is not cheap. Consequently, there is a need for accurate but cheaper approximation methods for EHVI. We propose and investigate a novel way of approximating the EHVI using Gauss-Hermite (GH) quadrature~\cite{jackel2005note,liu1994note}. In essence, GH approximates the integral of a function using a weighted sum resulting in fewer samples to approximate the EHVI.% Furthermore, we show that GH allows  considering correlations among objective functions \tc{TC: so does any sampling. They just need the distribution}, something that is currently possible to do with Monte Carlo approximations only. 

The rest of the article is structured as follows. In Section \ref{sect:background}, we  briefly describe multivariate predictive densities and EHVI, and then introduce the GH method in Section \ref{sect:GH}. In Section \ref{sect:exp}, we show the potential of the proposed idea of using GH by comparing it with analytical and MC approximations (for 2-3 objectives).
% with an exact calculation (for 2-3 objectives) and MC estimation. % \tc{TC: we just said closed form is available for any number of objectives} 
% and Monte Carlo (for 2-10 objectives) computations of EHVI. %\sout{Moreover, we embed the approximation in multi-objective BO (with EHVI as the acquisition function).}\mnote{RA: This sentence needs to be removed if these experiments do not finish in time.} 
Finally, conclusions are drawn in Section \ref{sect:concl}.

% we conclude and discuss future research directions in Section \ref{sect:concl}.

\section{Background}
\label{sect:background}
For multi-objective optimisation problems with $m$ objective functions to be minimised, given two vectors $\mathbf{z}$ and $\mathbf{y}$ in the objective space, we say that 
% a vector 
$\mathbf{z}$ dominates 
% another vector 
$\mathbf{y}$ if $z_i \le y_i$ for all $i=1, \dots , m$ and $z_j < y_j$ for at least one index $j$. A solution is Pareto optimal if no feasible solution dominates it. The set of Pareto optimal solutions in the objective space is called a Pareto front.

In multi-objective BO, the predictive distribution due to a solution with independent models is defined as:
\begin{align*}
    \mathbf{y} \sim \mathcal{N}(\bm{\mu},\diag(\sigma^2_1, \dots, \sigma^2_m)), 
    % \prod_{i=1}^m\mathcal{N}(\mu_i,\sigma_i^2),
\end{align*}
% \vspace{-1em}
where $m$ is the number of objectives  and $\bm{\mu} = (\mu_1, \dots, \mu_m)^\top$ is the mean vector, with $\mu_i$ and $\sigma_i$ being the mean and standard deviations of the predictive density for the $i^{th}$ objective. To quantify the correlation between objectives, a multi-task surrogate model can be used. The distribution of a solution with a single multi-task model is defined with a multi-variate Gaussian distribution:
\begin{align*}
    \mathbf{y} \sim \mathcal{N}(\bm{\mu},\Sigma),
\end{align*}
where $\bm{\mu}$ is the vector of means and $\Sigma$ is the covariance matrix that quantifies the correlation between different objectives. It should be noted that considering only the diagonal elements of $\Sigma$ would ignore any correlations between objectives, and result in an independent multivariate predictive density.

%\tc{TC: How about Background instead of Fundamentals?}
%To set the foundation for the rest of this paper, 
% We now provide a brief recap of BO %\sout{an established methodology for solving expensive optimization problems}, and 
% and the EHVI. %\sout{which has become a popular acquisition function to drive BO in a multi-objective setting.} \tc{TC: Repetitive}   
 
% \subsection{Bayesian optimisation}
% \input{BO}

% \subsection{Approximating expected hypervolume improvement}

The hypervolume measure~\cite{zitzler2003performance} is a popular indicator to assess the quality of a set of solutions to a multi-objective optimisation problem. Thus it is often used to compare %determine the performance of
multi-objective optimisation algorithms or for driving the search of indicator-based multi-objective optimisation algorithms. %Consequently, in the case of expensive function evaluations, the EHVI~\cite{emmerich2005single} has become popular. %a natural extension of EI --- a popular acquisition function for the single-objective case --- to the multi-objective case. 
The EHVI answers the question of what the expected improvement of the hypervolume is if some new candidate solution $\mathbf{x}$ would be added to an existing set of solutions. Consequently, the solution with the highest EHVI may be the one worth an expensive function evaluation. %that would be evaluated using a real expensive function. 
To avoid ambiguity, in the following, we provide formal definitions of the concepts discussed here, before discussing methods to calculate the EHVI.

%Emmerich et al. \cite{} extended the single-objective EI to multi-objective expected hypervolume improvement (EHVI). There is a closed form expression for EHVI for two and three objectives when the models built for each objective function independently. For more than three objectives, one needs to rely on approximations e.g.\ using Monte Carlo. Recently, a closed form expression of EHVI was proposed in \cite{Daulton} for any number of objectives. The EHVI can be used after building multi-task $\mGP$. The multi-task model can quantify the correlation between diffrent objectives and can help in improving the convergence \cite{}. However, there is no closed form expression for EHVI when the objectives are correlated. Therefore, one needs to rely on the Monte Carlo approximations.  \todo[inline]{TC:If there is closed form for any number of objectives, why we need Monte Carlo or even Gauss Hermite? What is the justification we wrote in the paper? Ahh for correlated objectives, I found the answer. But are we doing some tests for correlated objectives?}. In this work, we built independent models for each objective function and use EHVI as the acquisition function. The main steps of a multi-objective BO are summarised in Algorithm \ref{alg:BO}.

% We also built a multi-task $\mGP$ which quantifies the correlation between different objective functions. The correlation can be helpful in finding the conflict between different objectives and can improve the convergence \cite{}. One of the main disadvantages of using multi-task $\mGP$ is that there is no closed for expression for EHVI. 

\begin{definition}[Hypervolume indicator]
Given a finite set of $k$ points (candidate solutions) $P=\{\mathbf{p}_1,\dots,\mathbf{p}_k\} \subset \mathbf{R}^m$ where $\mathbf{p}_i = (f_1(\mathbf{x}_i), \dots, f_m(\mathbf{x}_i) ))^\top$ for an optimisation problem with $m$ objectives, the hypervolume indicator (HI) of $P$ is defined as the Lebesgue measure of the subspace (in the objective space) dominated by $P$ and a user-defined reference point $\mathbf{r}$~\cite{yang2015expected}:
\begin{equation*}
    HI(P) = \Lambda (\cup_{\mathbf{p}\in P}[\mathbf{p},\mathbf{r}]),
\end{equation*}
where $\Lambda$ is the Lebesgue measure on $\mathbf{R}^m$, and $\mathbf{r}$ chosen such that it is dominated by all points in $P$, and ideally also by all points of the Pareto front. 
\end{definition}

\begin{definition}[Hypervolume contribution]
Given a point $\mathbf{p}\in P$, its hypervolume contribution with respect to $P$ is $\Delta HI(P,\mathbf{p}) = HI(P) - HI(P\backslash\{\mathbf{p}\})$.
\end{definition}

\begin{definition}[Hypervolume improvement]\label{def:imp}
Given a point $\mathbf{p}\notin P$, its hypervolume improvement with respect to $P$ is $I(\mathbf{p}, P) = HI(P \cup \{\mathbf{p}\}) - HI(P)$.
\end{definition}

\begin{definition}[Expected hypervolume improvement]
Given a point $\mathbf{p}\notin P$, its expected hypervolume improvement (EHVI) with respect to $P$ is 
\begin{equation*}
\label{eq:EHVI}
    \int_{\mathbf{p}\in\mathbb{R}^m} HI(P,\mathbf{p})\cdot PDF(\mathbf{p})d\mathbf{p}, 
\end{equation*}
where $PDF(\mathbf{p})$ is the predictive distribution function of $\mathbf{p}$ over points in the objective space.
\end{definition}

The EHVI can be computed analytically for any number of objectives assuming they are uncorrelated, but this requires partitioning the objective space, which can be prohibitively expensive for $m>3$ objectives. Consequently, there is considerable interest in finding %investigating 
more efficient ways to compute EHVI, see e.g.~\cite{couckuyt2014fast,daulton2020differentiable,emmerich2016multicriteria,emmerich2011hypervolume,hupkens2015faster,yang2017computing}. MC integration is an alternative to an exact computation of EHVI. It is easy to use in practice and has been the method of choice for problems with $m>3$ objectives. Given a multivariate Gaussian distribution from which samples are drawn, or $\mathbf{p}_i \sim \mathcal{N}(\bm{\mu}, \Sigma)$, %with mean $\bm{\mu}$ = (\mu_1, \dots, \mu_m)^\top$ is the mean vector and covariance matrix $\Sigma$, 
and a set of points $P$ (e.g. an approximation of the Pareto front), then the MC approximation of EHVI across $c$ samples is
\begin{equation}
\label{eq:mc}
    \frac{1}{c} \sum_{i=1}^c I(\mathbf{p}_i, P), 
\end{equation}
where $I(\mathbf{p}_i, P)$ is the hypervolume improvement (see Definition~\ref{def:imp}). The approximation error is given by $e = \sigma_\text{M}/\sqrt{c}$, where $\sigma_\text{M}$ is the sample standard deviation \cite{koehler2009assessment}. Clearly, as the sample size $c$ increases, the approximation error reduces, namely in proportion to $1/\sqrt{c}$. In other words, a hundred times more samples will result in improving the accuracy by ten times.

\begin{comment}
The EHVI can be calculated exactly  %EHVI exactly is limited to the case of $m=2$~\cite{emmerich2011hypervolume,hupkens2015faster,emmerich2016multicriteria} and $m=3$~\cite{hupkens2015faster,couckuyt2014fast,yang2017computing}. 
with recent research trying to make this calculation ever so faster (see, for example,~\cite{hupkens2015faster,emmerich2016multicriteria,yang2017computing,yang2019efficient}). 
%, and is associated with a runtime complexity in $O(n^2)$ and $O(n^3)$, respectively, where $n$ is the number of non-dominated points in $P$.
MC integration is an alternative to an exact computation of EHVI. Given a multivariate predictive distribution $\mathcal{N}(\bm{\mu}, \Sigma)$, where $\bm{\mu} = (\mu_1, \dots, \mu_m)^\top$ is the mean vector and $\Sigma$ is the covariance matrix, we can take a sample $\mathbf{p} \sim \mathcal{N}(\bm{\mu}, \Sigma)$. For $i$th sample and a reference front $P$, we can calculate the improvement $I(\mathbf{p}^i, P)$. For $k$ samples, the MC approximation of the EHVI is: 
\begin{align}
\label{eq:mc}
    \frac{1}{k} \sum_{i=1}^k I(\mathbf{p}^i, P).
\end{align}

The overall approximation error of the MC approach is given by: $e = \frac{\sigma_M}{\sqrt{c}}$, where $\sigma_M$ is the sample standard deviation \cite{koehler2009assessment}. Clearly, with a large number of samples, we reduce the error in proportion to $1/\sqrt{k}$. In other words, a hundred times more samples will result in improving the accuracy of the approximaion by ten times. 
\end{comment}

Typically, evaluating the improvement due to a single sample can be rapid. Even if we consider a large $c$, it is often not that time-consuming to compute the EHVI for a single predictive density. However, when we are optimising EHVI to locate the distribution that is the most promising in improving the current approximation of the front, an MC approach may become prohibitively expensive with a large enough $c$ for an acceptable error level. Therefore, alternative approximation methods that are less computationally intensive are of interest. In the next section, we discuss such an approach based on GH quadrature.

\section{Gauss-Hermite approximation}
\label{sect:GH}
The idea of GH approximation is based on the concept of Gaussian quadratures, which implies that if a function $f$ can be approximated well by a polynomial of order $2n-1$ or less, then a quadrature with $n$ nodes suffices for a good approximation of a (potentially intractable) integral~\cite{jackel2005note,liu1994note}, i.e. 
\begin{equation*}
    \int_{D} f(\mathbf{x}) \psi(\mathbf{x}) \,dx \; \approx\; \sum_{i=1}^{n}w_i f(\mathbf{x}_i),
\end{equation*}
where $D$ is the domain over which $f(\mathbf{x})$ is defined, and $\psi$ a known weighting kernel (or probability density function). The domain $D$ and weighing kernel $\psi$ define a set of $n$ weighted nodes $\mathcal{S}=\{\mathbf{x}_i, w_i\}, i=1,\dots,n$, where $\mathbf{x}_i$ is the $i$th deterministically chosen node and $w_i$ its associated quadrature weight. We refer to this concept as \textit{Gauss-Hermite} if $D$ is infinite, i.e., $D\in(-\infty,\infty)$, and the weighting kernel $\psi$ is given by the density of a standard Gaussian distribution. 

The location of the nodes $\mathbf{x}_i$ are determined using the roots of the polynomial of order $n$, % in the sequence of orthonormal polynomials $\{\pi_j\}$ generated according to $\langle \pi_j|\pi_k\rangle := \int_{D} \pi_j(x)\pi_k(x)\psi(x) \,dx  = \delta_{jk}$, 
while the weights $w_i$ are computed from a linear system upon computing the roots~\cite{jackel2005note}; the interested reader is referred to~\cite{teukolsky1992numerical} for technical details of this calculation. Intuitively, one can think of the selected nodes as representatives of the Gaussian distribution with the weights ensuring convergence as $n$ increases and a low approximation error for a given $n$~\cite{elvira2019gauss}. 

\begin{figure*}[t]
     \centering
     \begin{subfigure}[b]{0.32\textwidth}
         \centering
         \includegraphics[width=1\textwidth]{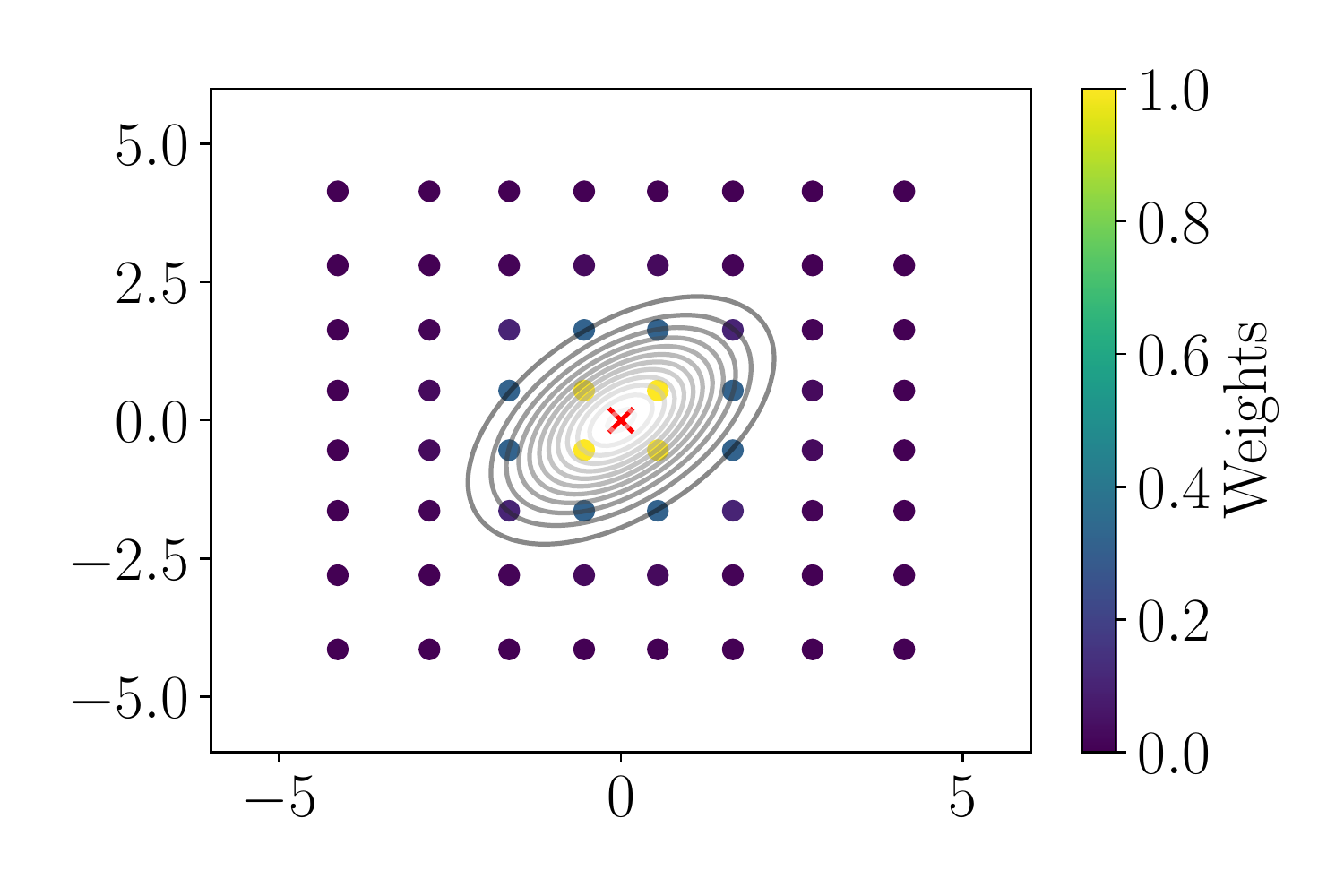}
         \caption{Sampling}
         \label{fig:gh_start}
     \end{subfigure}
     \hfill
     \begin{subfigure}[b]{0.32\textwidth}
         \centering
         \includegraphics[width=1\textwidth]{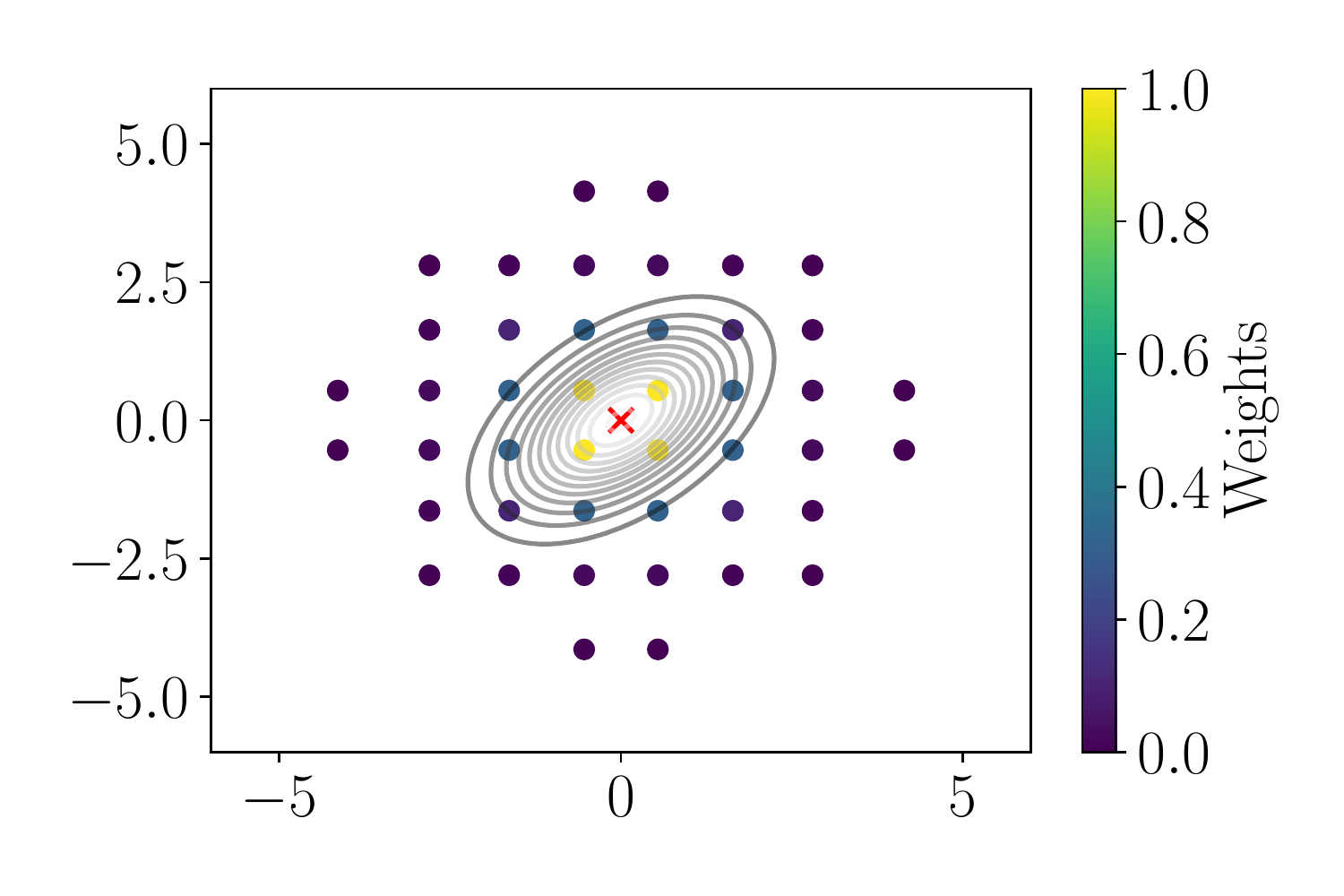}
         \caption{Pruning}
         \label{fig:gh_rotate}
     \end{subfigure}
     \hfill
     \begin{subfigure}[b]{0.32\textwidth}
         \centering
         \includegraphics[width=1\textwidth]{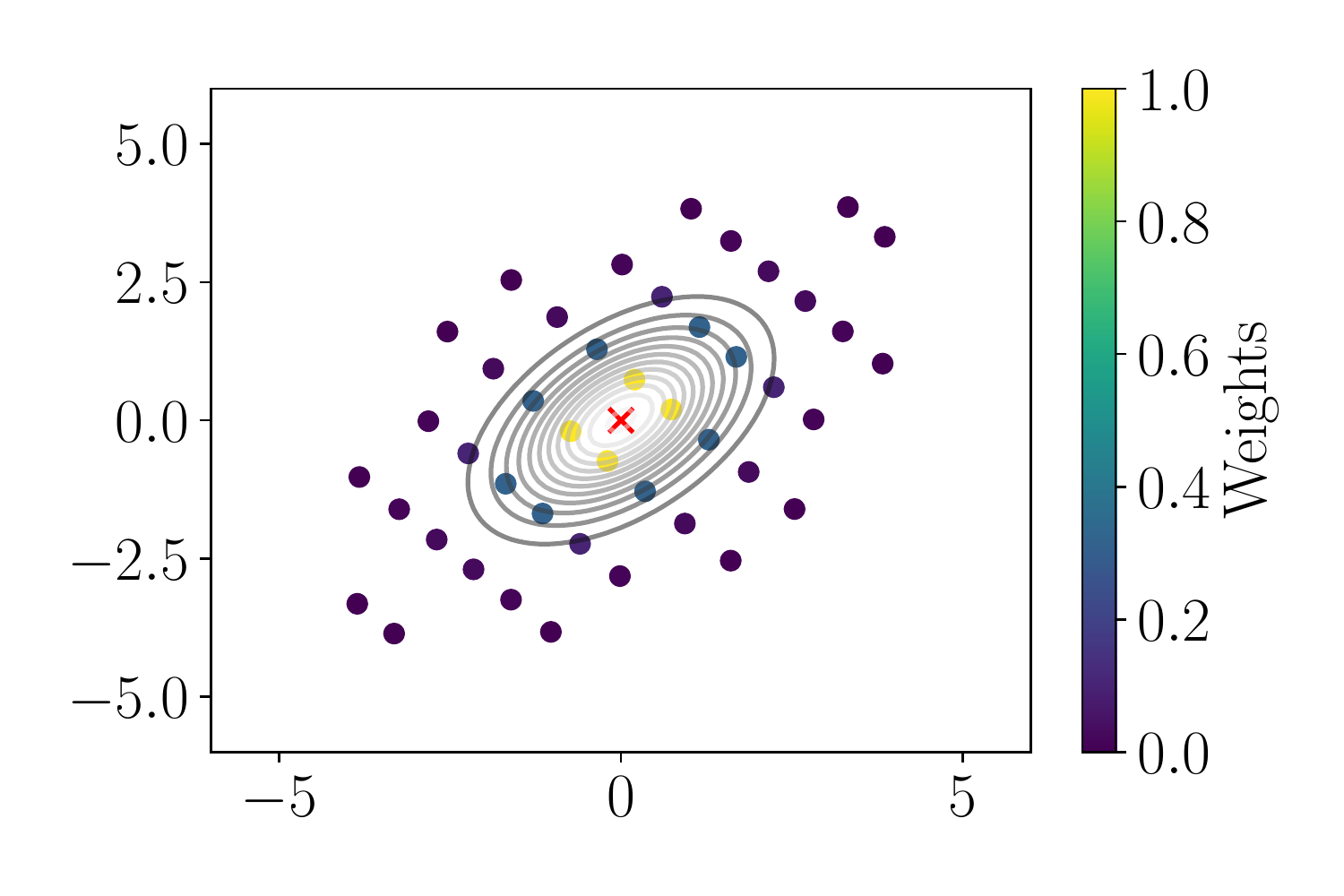}
         \caption{Rotation}
         \label{fig:gh_prune}
     \end{subfigure}
        \caption{An illustration of the process of generating the nodes and the associated weights using the GH quadrature for a two-dimensional ($m=2$) Gaussian density with the mean vector $\bm{\mu} = (0, 0)^\top$ and the covariance matrix $\Sigma = \begin{pmatrix}
  1 & 0.5 \\
  0.5 & 1 
\end{pmatrix}$. The parameters used are:\ $n=8$ points per dimension and a pruning rate of $r=0.2$ (i.e., 20\%). 
The dots represent the nodes, and the colours represent the respective weights.
The contours show the Gaussian density  
% The weights are proportional to the probability density of the Gaussian distribution, contours of which are depicted in gray 
with the outermost contour corresponding to two-standard deviation. 
}
        \label{fig:gauss-hermite}
\end{figure*}

Extending the one-dimensional GH 
%Gauss-Hermite 
quadrature calculations to multivariate integrals is achieved by expanding the one-dimensional grid of nodes to form an $m$-dimensional grid, which is then pruned, rotated, scaled, and, finally, the nodes are translated. % and covariance matrix of the multivariate Gaussian distribution over which the integral is calculated. 
Figure~\ref{fig:gauss-hermite} illustrates the key steps of this process for a two-dimensional ($m=2$) integral. The weights of the $m$-variate quadrature points are the product of the corresponding $m$ one-dimensional weights; for $m=2$, this leads to the following two-dimensional Gaussian quadrature approximation:
\begin{equation*}
    \sum_{i=1,j=1}^{n,n}w_iw_j f(\mathbf{x}_i,\mathbf{x}_j).
\end{equation*}
The pruning step eliminates nodes that are associated with a low weight (i.e., points on the diagonal as they are further away from the origin); such nodes do not contribute significantly to the total integral value, hence eliminating them improves computational efficiency. Rotating, scaling and translating nodes account for correlations across dimensions, which is often present in practice. 
The rotation and scaling are conducted using a rotation matrix constructed from the dot product of the eigenvector and the eigenvalues of the covariance matrix, and the translation is performed by adding the mean vector.

% This can be achieved using, for example, an approach based on a Cholesky decomposition or decompose spectrally~\cite{jackel2005note}; the reader is referred to~\cite{jackel2005note} for more details. Here we use correlation incorporation by the aid of spectral decomposition\mnote{RA: Alma, I do not remember if you actually do rotation. Please update this sentence with the correct approach (Cholesky vs spectrally) or say that we do not do rotation and why not}.

%While the area around one-dimensional Gauss-Hermite quadrature calculations has been studies widely with many software libraries available to carry out the calculations, there is relatively little work on extending this concept to the multi-dimensional case. 

%Somwhere to include Monte Carlo vs Quadrature methods: MC integration is a stochastic method, but its computation can be prohibitively expensive, especially when the integral is computed many times. Quadrature methods are deterministic rather than stochastic, and can be less computationally expensive, especially for lower-dimension integrals. 

Note that this approach may result in a combinatorial explosion of nodes in approximating a high-dimensional multivariate Gaussian distribution. Given $n$ nodes per dimension for an $m$-dimensional space, the total number of nodes generated is $K = \lfloor n^m (1- r)\rfloor$, where $r\in[0,1]$ is the pruning rate (the greater $r$, the more nodes are discarded with $r=0$ corresponding to not discarding any nodes). For instance, using $n=5$, $m=10$ and $r=0.2$, we have $K=7812500$ nodes. Clearly, this would be computationally more expensive than MC approximation. Therefore, for high-dimensional integration the default GH approach may not be suitable. As a rule of thumb, we propose that if the number of nodes from GH goes beyond the number of samples required for a good MC approximation, then one should use the latter, instead. 

It should be noted that there is some work on high-dimensional GH approximations, e.g.~\cite{holtz2008sparse}, but we do not investigate these in this paper. 
% \todo[inline]{AR: Here we should have a discussion about how the overall number of points changes with the dimensionality of the integrals, and the practical guidance on where $G$ remains competitive with respect to $M$.}

\subsection{Gauss-Hermite for approximating EHVI}\label{algorithmGH}

\begin{figure}[t]
    \centering
    \includegraphics[width=0.6\textwidth, trim={80mm 0mm 80mm 0mm}, clip=true]{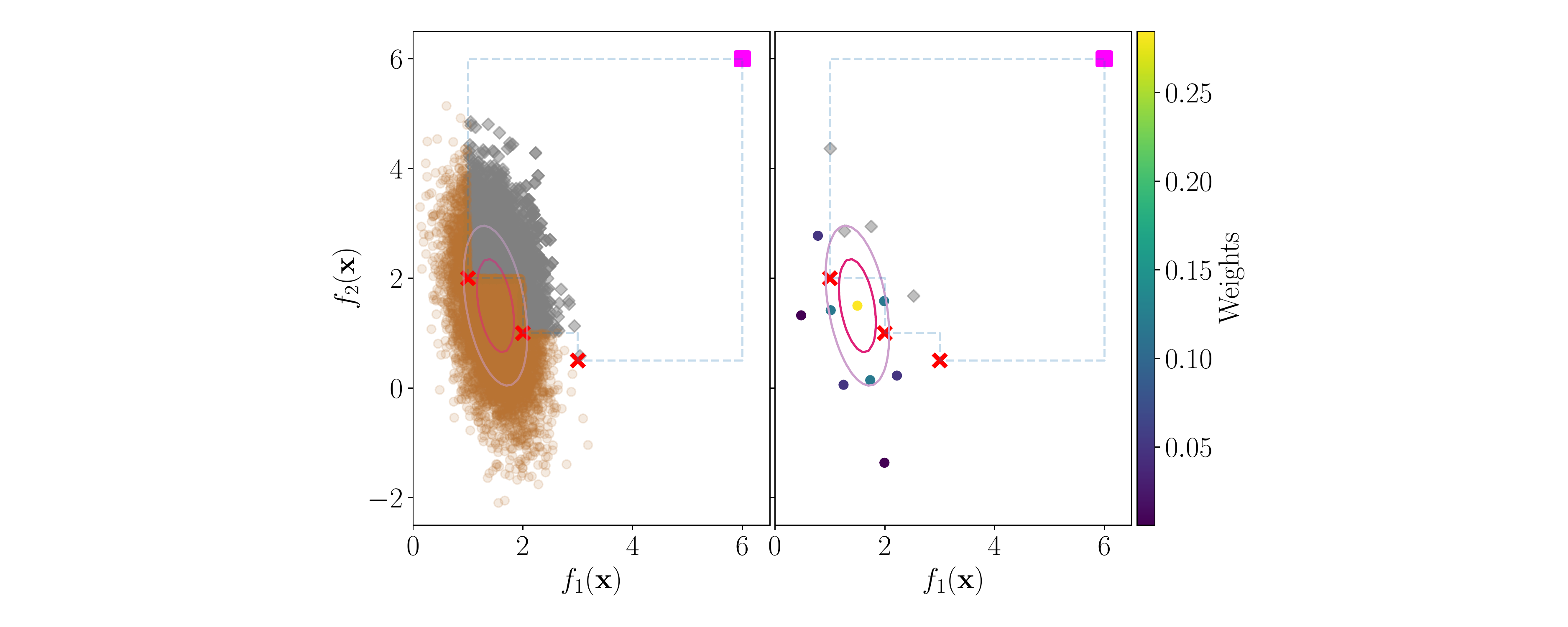}
    \caption{A visual comparison of MC (\textit{left}) and GH (\textit{right}) samples in the objective space (assuming a minimisation problem). The approximation of the Pareto front is depicted with red crosses, and the reference vector $\mathbf{r}$ for hypervolume computation is shown with a magenta square. The blue dashed line outlines the dominated area. We used a two-variate Gaussian distribution to generate samples with the mean vector $\bm{\mu} = (1.5, 1.5)^\top$ and the covariance matrix $\Sigma = \begin{pmatrix}
 0.16 & -0.15\\
 -0.15 &  1.01 
\end{pmatrix}$; the pick contours represent this density.
% The pink contours represent the Gaussian predictive density. 
The gray diamonds are dominated by the approximated front.
% , and therefore do not contribute to the calculation of EHVI.
}
    \label{fig:mc_gh_samples}
\end{figure}

To approximate the EHVI, we use $K$ samples (nodes) and associated weights from GH quadrature as follows: 
\begin{align}
\label{eq:gh}
    \sum_{i=1}^{K} \omega_i I(\mathbf{p}_i, P),
\end{align}
where $P$ is the approximated Pareto front, $\mathbf{p}_i$ is the $i$th sample, and $\omega_i = \prod_{j=1}^m w_j(\mathbf{x}_i)$ is the weight in an $m$-dimensional objective space corresponding to the sample $\mathbf{x}_i$. This is effectively a weighted sum of the contributions, where the weights vary according to the probability density. This is also illustrated in the \textit{right panel} of Figure~\ref{fig:mc_gh_samples}: The dots show the GH samples (nodes). The grid of points covers an area that is consistent with the underlying Gaussian distribution. Since we know how the probability density varies, we can generate proportional weights, which in turn permits us to derive a good approximation with only a few points in the grid.  

On the other hand, with MC in \eqref{eq:mc}, every sample (dots in Figure \ref{fig:mc_gh_samples}, \textit{left panel}) contributes equally to the average EHVI. 
% The samples (dots in Figure \ref{fig:mc_gh_samples}, \textit{left panel}) contribute equally towards the average improvement. 
Hence, a sample is somewhat unrelated to the intensity of the underlying probability density at that location. As such, with few samples, we may not derive a good approximation. It should be noted that the gray diamonds (in both panels) 
% \mnote{RA: the left panel has only gray samples, so this statement is wrong: Using a different symbol eg diamond, instead of a grey dot, may be better.} 
are dominated by the approximation of the Pareto front, and therefore there is no improvement (see Definition~\ref{def:imp}) due to these solutions. Hence, these gray diamonds do not contribute to the EHVI for either of the methods.

% In calculating EHVI, the dominated points (shown in gray) do not contribute. In $M$ (\textit{left}), for each of the darker non-dominated points we compute the hypervolume contribution over the current Pareto front and take the average. In $G$, the contribution of each  

\section{Experimental study}
\label{sect:exp}
%To answer the question whether Gauss-Hermite is a valid approach for approximating EHVI, we perform an experimental study (Section~\ref{ExAnalysis}) that investigates how accurate an integral can be estimated with the (pruned) set of nodes in comparison to using Monte Carlo sampling, with a particular focus on correlated multivariate distributions. %(i)~how accurate an integral can be estimated with the (pruned) set of nodes in comparison to using Monte Carlo sampling with the same number of points (or more points), (ii)~the impact of pruning---a key parameter in the Gauss-Hermite method---on approximation performance, and (iii)~the suitability of Gauss-Hermite when used in EHVI-EGO to approximate the expected hypervolume improvement (i.e. the infill criteria of EHVI-EGO). 
%Before we present our results and analysis, we provide the experimental setup in the next section. 

%\subsection{Experimental setup}

In this section, we focus on comparing the accuracy of GH and MC approximations with respect to the analytical calculation of EHVI ($A$) introduced in~\cite{couckuyt2014fast,emmerich2011hypervolume}. As the analytical method is only suitable for independent multivariate Gaussian densities, we firstly investigate the efficacy of the approximation methods for uncorrelated densities for $m=2$ and $3$, and then expand our exploration to correlated multivariate densities. We use popular test problems:\ DTLZ 1-4, 7~\cite{deb2005scalable}, and WFG 1-9~\cite{huband2006review}. They were chosen as they allow us to validate the efficacy of the approximation methods for Pareto fronts with diverse features;\ e.g., DTLZ2 and WFG4 have concave, DTLZ7 and WFG2 have disconnected, and DTLZ1 and WFG3 have linear Pareto fronts.

Our strategy was to first generate a random multivariate distribution, and then, for an approximation of the known Pareto front, compute the EHVI due to this random distribution analytically and with the two approximation methods (GH and MC). Using this approach, we aimed to collect data on a range of randomly generated multivariate distributions and inspect the agreement between analytical measurements and approximations. To quantify this, we used Kendall's $\tau$ rank correlation test~\cite{abdi2007kendall}, which varies between $[-1,1]$ with $1$ showing perfectly (positively) correlated ordering of the data by 
% the three 
a pair of competing 
% EHVI 
methods. The test also permits the estimation of a $p$-value, which, if below a predefined level $\alpha$ indicates that results are significant. In this paper, we set $\alpha = 0.05$, however, in all cases, we found the $p$-value to be practically zero, hence indicating significance in the results. 

To implement the GH approximation, we converted existing R code\footnote{\url{https://biostatmatt.com/archives/2754}} into Python; our code is available to download at~\url{github.com/AlmaRahat/EHVI\_Gauss-Hermite}. If not stated otherwise, MC uses $10,000$ samples, and GH uses a pruning rate of $r=0.2$. For GH, we investigate different numbers of nodes (points) $n$ per dimension, and use the notation $GH_{n}$ to indicate this number. Any results reported are 
% average 
results obtained across $100$ randomly generated multivariate Gaussian distributions to generate as many EHVIs.

\begin{comment}
\begin{table}[t]
    \centering
    \caption{Default parameter settings as used in the experimental study.\label{tab:params}}
\begin{tabular}{*{2}{c}}
\toprule
Parameter & Value \\
\midrule
Pruning fraction & $20\%$ \\
Monte Carlo trials & $10000$\\
\bottomrule
\end{tabular}
\end{table}
\end{comment}

\subsection{Uncorrelated multivariate Gaussian distribution}
To generate a random multivariate distribution, we first take a reference front $P$. We then calculate the maximum $p_{\max}^i$ and minimum $p_{\min}^i$ values along each objective function $f_i$. The span along the $i$th objective is thus $s = p_{\max}^i - p_{\min}^i$. Using this, we construct a hyper-rectangle $H$ which has lower and upper bounds at vectors $\mathbf{l} = (l_1, \dots, l_m)$ and $\mathbf{u} = (u_1, \dots, u_m)$, respectively, with $l_i = p_{\min}^i - 0.3s$  and $u_i =  p_{\max}^i + 0.3s$. We take a sample from $H$ uniformly at random to generate a mean vector $\bm{\mu}$. The covariance matrix must be a diagonal matrix with positive elements for an independent multivariate distribution. Hence, we generate the $i$th diagonal element by sampling uniformly at random in the range $[0, u_i - l_i]$.

\begin{figure*}[t]
     \centering
     \begin{subfigure}[b]{0.32\textwidth}
         \centering
         \includegraphics[width=1\textwidth]{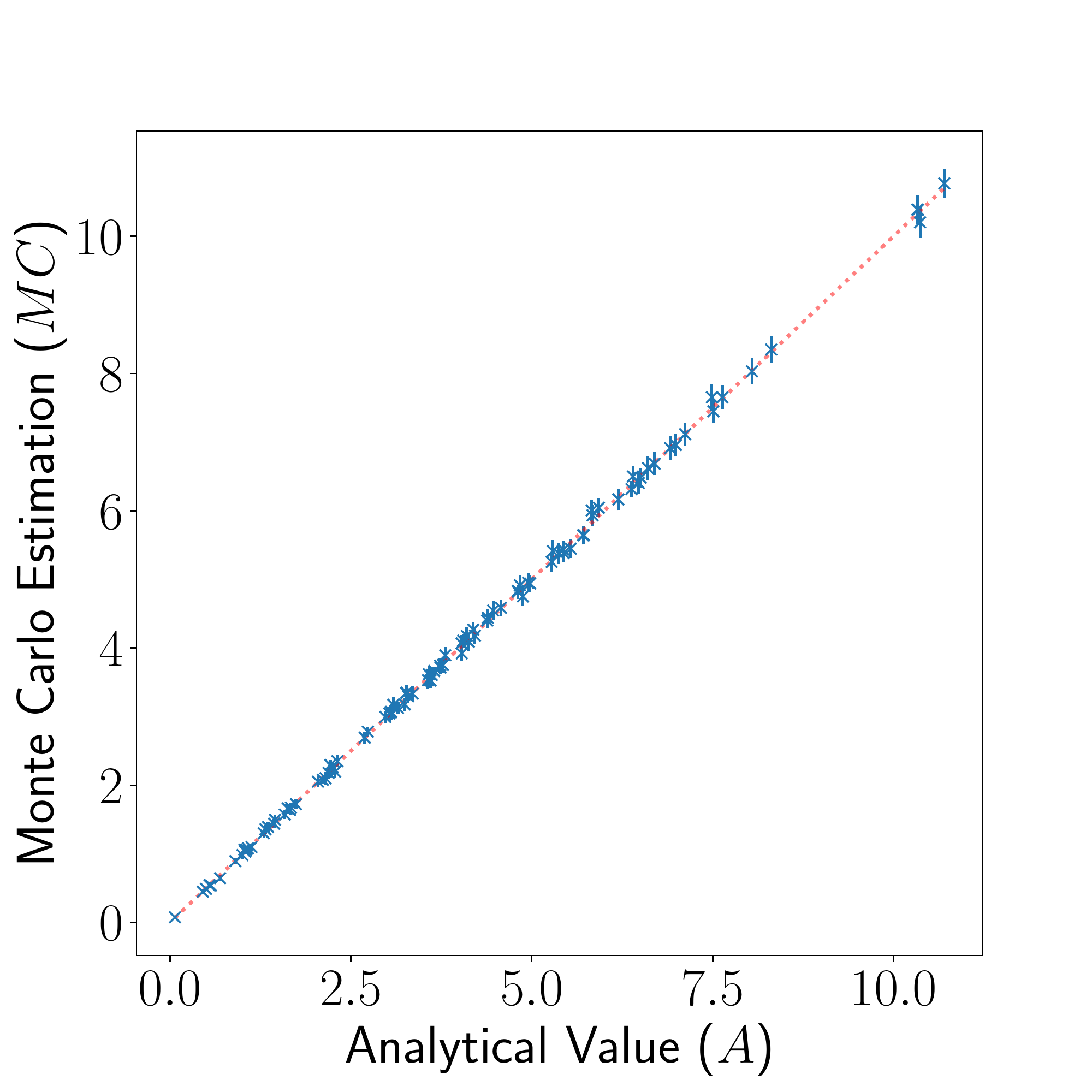}
         \caption{}
         \label{fig:mc_v_meas}
     \end{subfigure}
     \hfill
     \begin{subfigure}[b]{0.32\textwidth}
         \centering
         \includegraphics[width=1\textwidth]{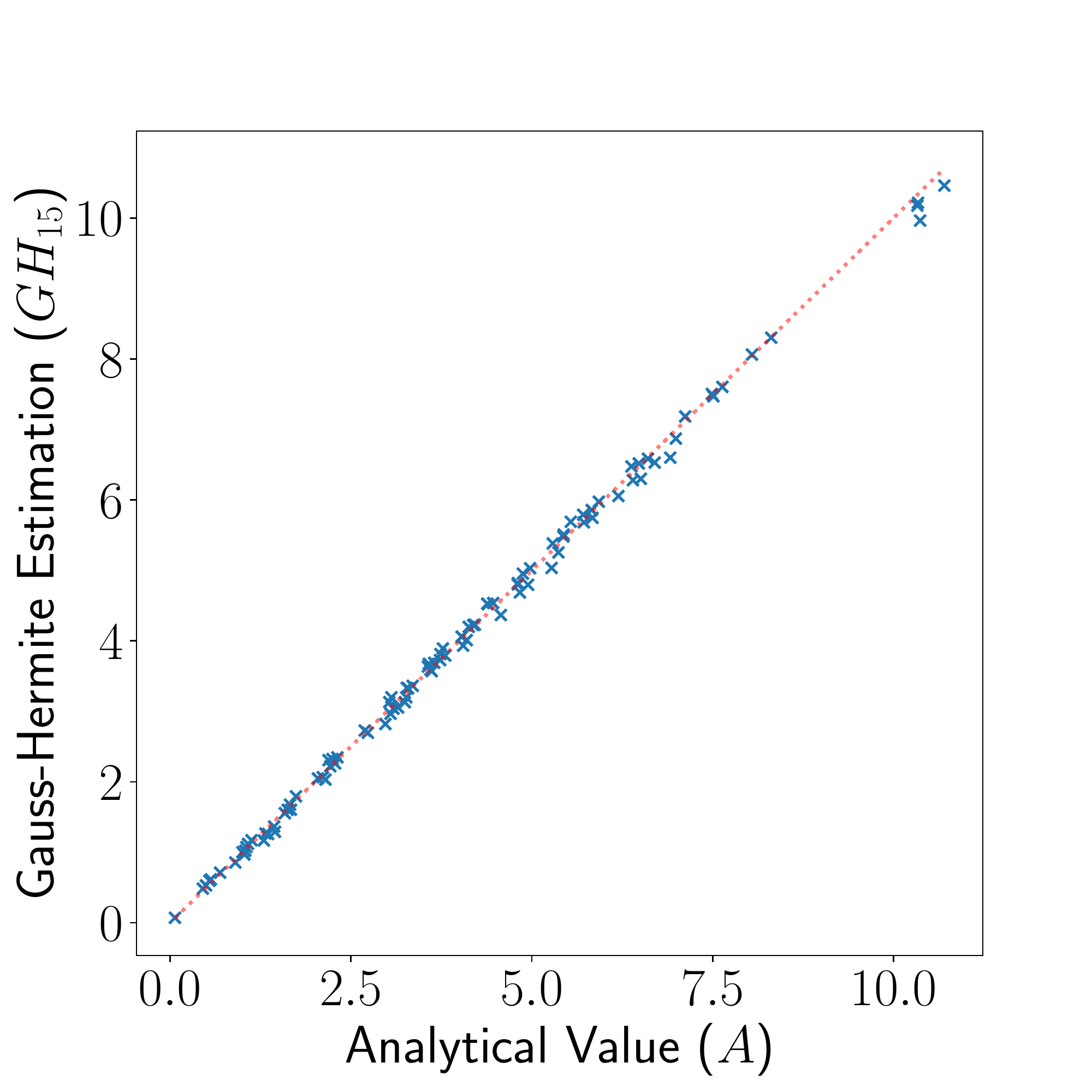}
         \caption{}
         \label{fig:gh_v_meas}
     \end{subfigure}
     \hfill
     \begin{subfigure}[b]{0.32\textwidth}
         \centering
         \includegraphics[width=1\textwidth]{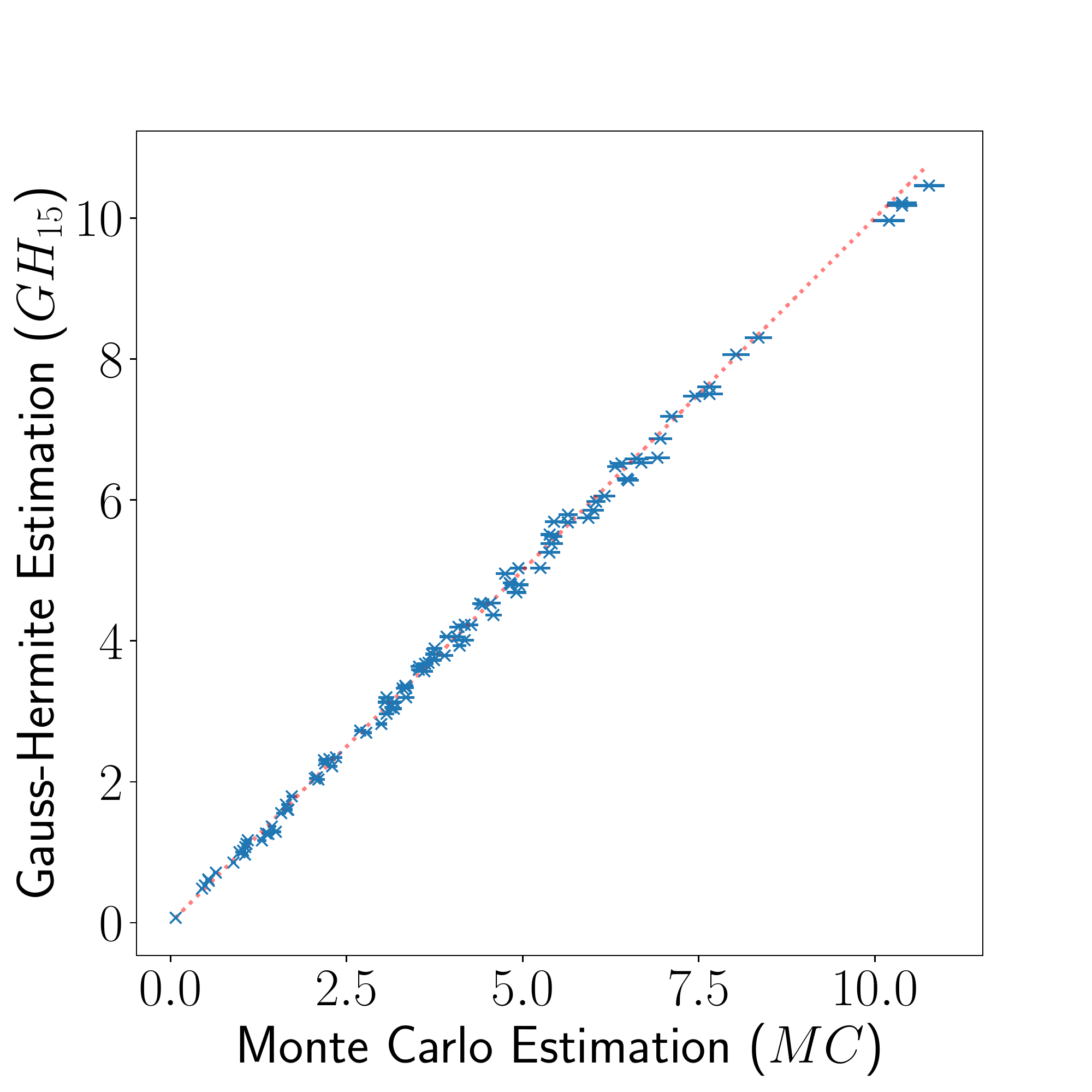}
         \caption{}
         \label{fig:gh_v_mc}
     \end{subfigure}
        \caption{Efficacy of MC and GH (with $15$ points per dimension, $GH_{15}$) approximations in comparison to analytical measurements of EHVI for the DTLZ2 problem with $2$ objectives and $100$ randomly generated multivariate Gaussian distributions. The dotted red-line depicts the performance of the perfect approximations. MC approximations used $10,000$ samples. % In \ref{fig:mc_v_meas} and \ref{fig:gh_v_mc} show the error estimates of the MC approximations with $10,000$ samples. 
        % The horizontal (in \ref{fig:mc_v_meas}) and vertical (in \ref{fig:gh_v_mc}) axes show the error estimates of the Monte Carlo approximations with $10,000$ samples. 
        In all cases, we observe strong rank correlations with Kendall's $\tau$ coefficient over $0.97$ with practically zero $p$-values.}
        \label{fig:m_g_a}
\end{figure*}

\begin{figure}[t]
\centering
\includegraphics[width=0.4\textwidth, trim={0 10mm 0 10mm}, clip=true]{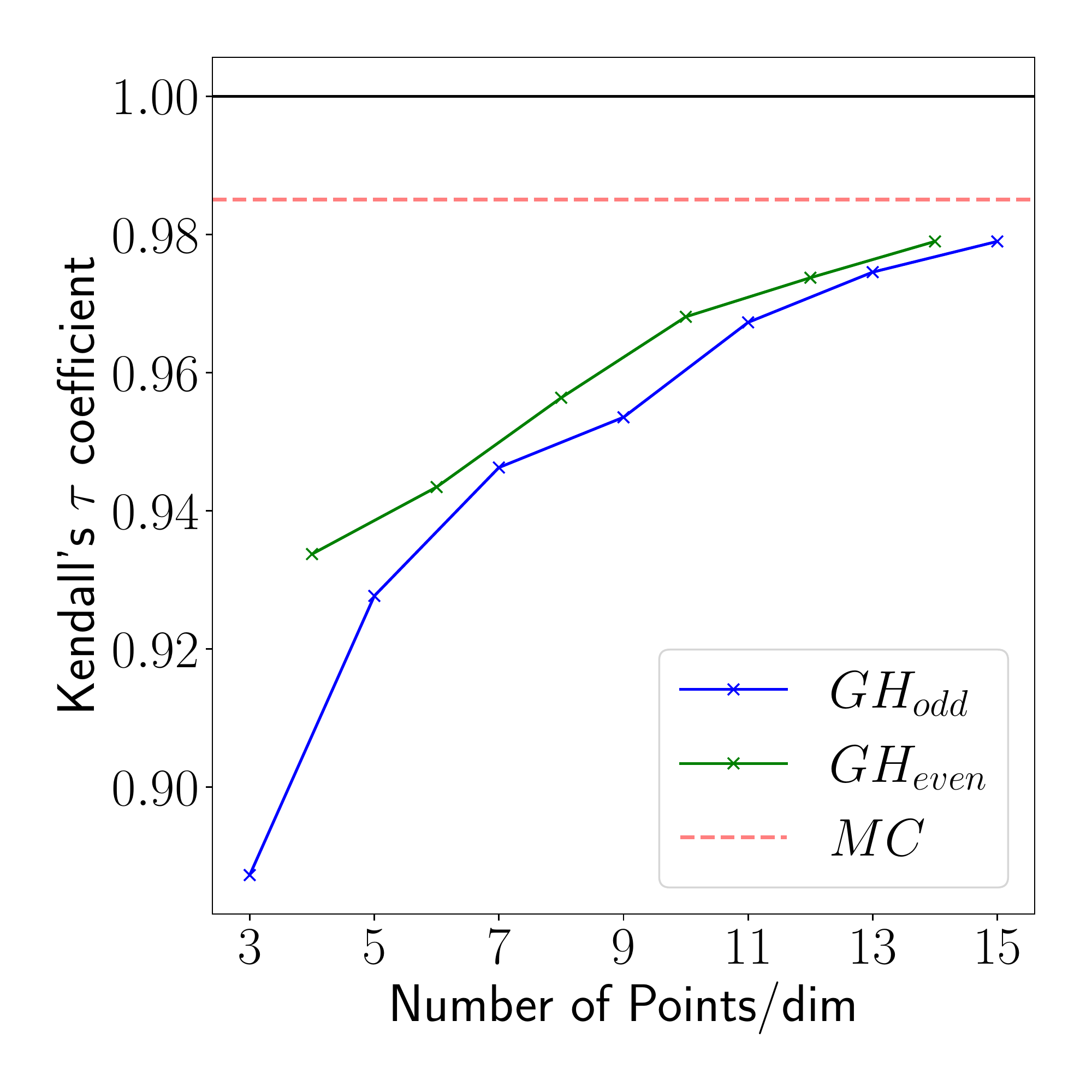}
\caption{Increase in accuracy with the increase in the number of points per dimension between $3$ to $15$ for the GH approximation with respect to the analytical result for DTLZ2 ($m=2$) and $100$ random multivariate distributions. The black horizontal line shows the theoretical upper bound for Kendall's $\tau$ coefficient. The red dashed horizontal line shows the coefficient for the EHVI using MC. The blue and red lines depict the increase in coefficient as we increase the number of points per dimensions for odd and even numbers, respectively, for GH. \label{fig:pts_per_dim}}
\end{figure}

\begin{figure}[t]
\centering
\includegraphics[width=0.7\textwidth, trim={0 5mm 0 5mm}, clip=true]{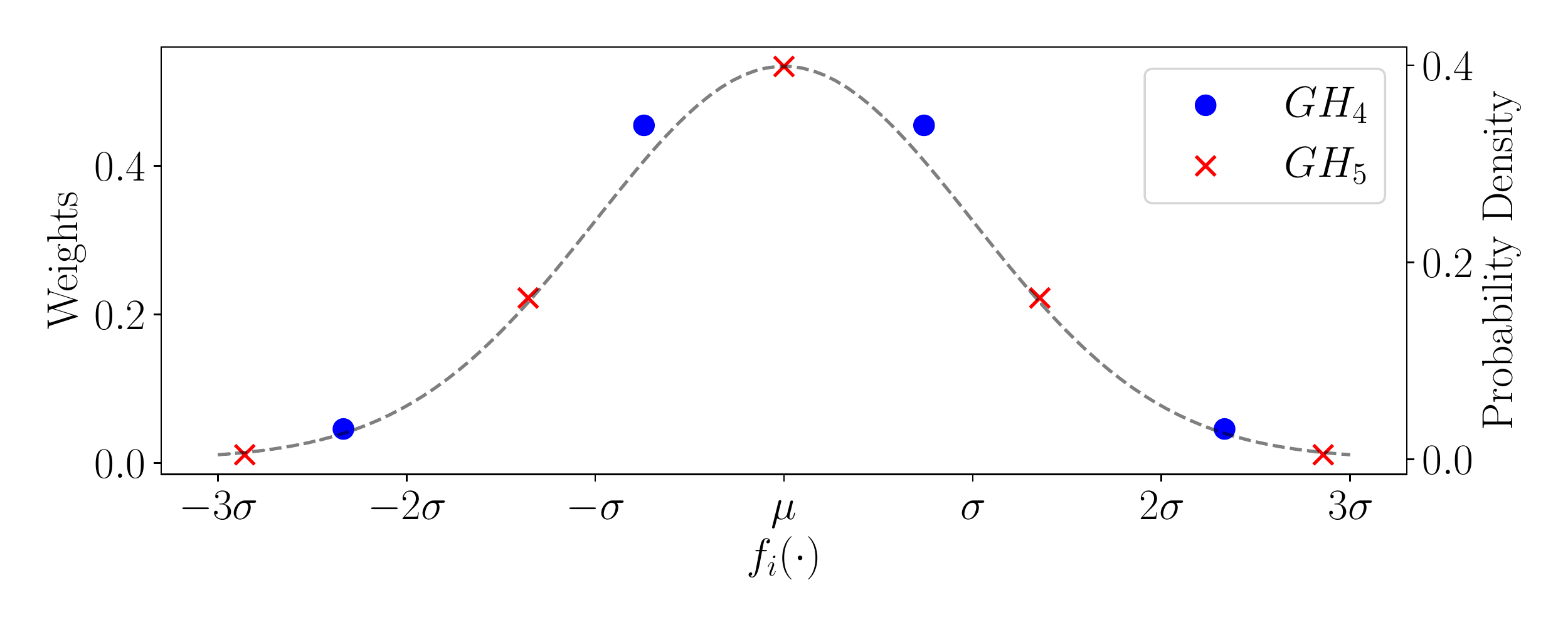}
\caption{An example of the distribution of GH nodes for $4$ ($GH_4$) and $5$ ($GH_5$) points per dimension (before pruning) for the standard Gaussian density with the mean $\mu = 0$ and the standard deviation $\sigma=1$ (shown in dashed black line). 
% With odd number of points $G$ would produce a node at  the mean $\mu$, whereas with even number of points per dimension a node at $\mu$ is missing. Inevitably, this will impact the estimation of EHVI. 
\label{fig:pts_dist}}
\end{figure}

Figure \ref{fig:m_g_a} shows an example comparison between the analytical (A), MC and GH computations of EHVI for DTLZ2. 
%, and $100$ randomly generated multivariate Gaussian distributions to generate as many expected improvements. For $G$ and $M$, we used the parameter settings in Table \ref{tab:params}. In addition, for $G$, we used $15$ points per dimension; the subscript indicates the number of points per dimension, i.e. $G_{15}$. 
The comparisons clearly show that the performances of MC and $GH_{15}$ are reliable with respect to $A$ with a Kendall's $\tau$ coefficient of over $0.97$ and associated $p$-value of (almost) zero. To investigate if there is an increase in accuracy with the number of points per dimension, we repeated the experiment by varying the number of points per dimension between $3$ and $15$ (see Figure \ref{fig:pts_per_dim} for results on the DTLZ2 problem with $m=2$). 
Interestingly, there is a difference between having an odd or an even number of points per dimension:\ there is often a dip in performance when we go from even to odd. In Figure \ref{fig:pts_per_dim}, we see that there is a slight decrease in the rank coefficient between $4$ and $5$ points per dimension. We attribute this decrease to how the points are distributed for odd and even numbers of points per dimension. When we have an odd number of points for GH, it produces a node at the mean of the distribution. If there is an even number of points per dimensions, there is no node at the mean (see Figure \ref{fig:pts_dist}). Because of this, the approximation may vary between odd and even number of points. Nonetheless, the monotonicity in accuracy improvement is preserved when the number of points is increased by two.

\begin{figure}[t]
    \centering
    \includegraphics[width=0.9\textwidth, trim={35mm 10mm 40mm 0}, clip=true]{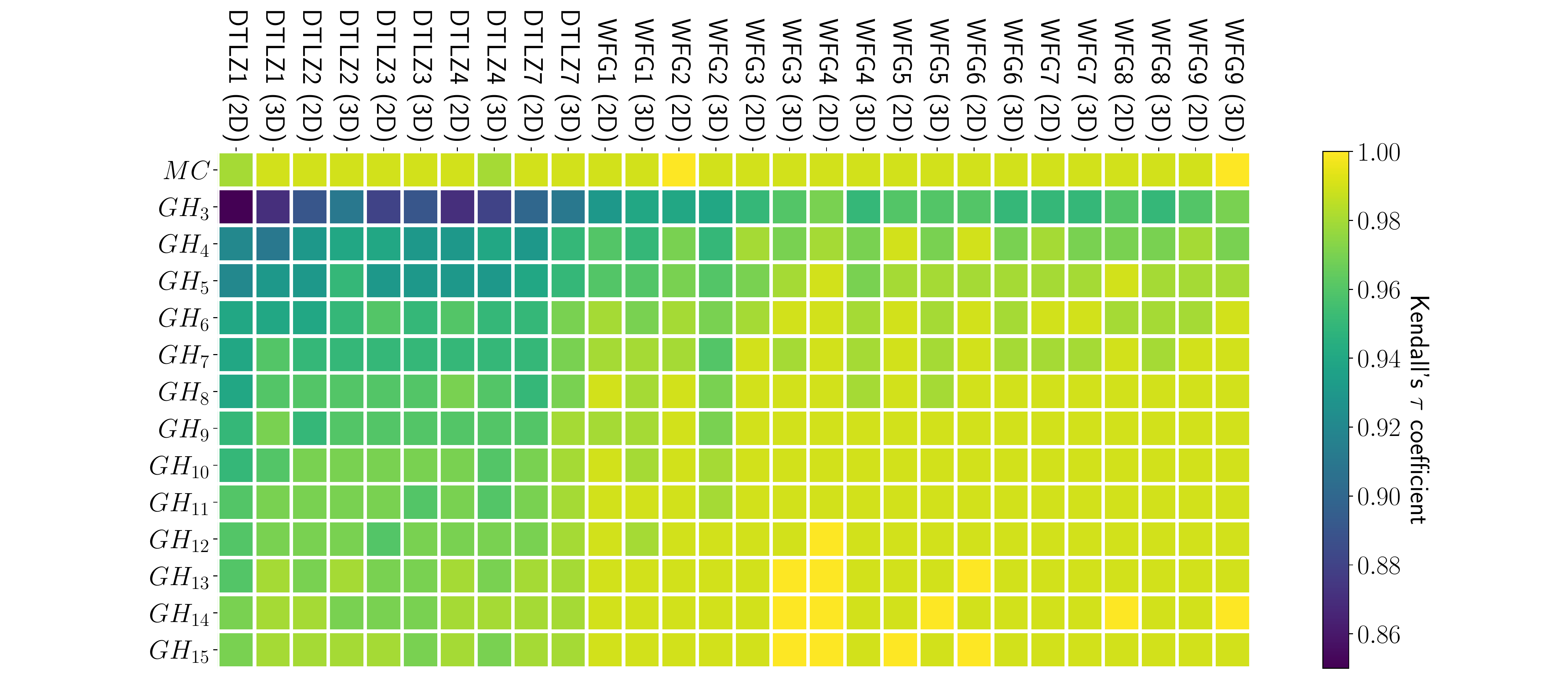}
    \caption{Performance comparison of MC and GH ($GH_n$, where $n$ is the number of points per dimension) with respect to the analytical EHVI for a range of test problems with $100$ randomly generated multivariate Gaussian distributions in each instance. Lighter colours correspond to better coefficient values. %The minimum coefficient value was just over $0.85$ for $GH_3$ on DTLZ1 (3D); this is a strong positive rank correlation with few points per dimension. 
    %Performance comparison of MC and GH ($GH_k$, where $k$ is the number of points per dimension) approximations with respect to the analytical EHVI for a range of test problems (along the horizontal axis) with $100$ randomly generated multivariate Gaussian distributions in each instance. The lighter colours corresponds to better coefficient values. The minimum coefficient value was just over $0.85$ for $GH_3$ on DTLZ1 (with $m=2$); this is a strong positive rank correlation with few points per dimension. There is a clear trend of increasing the accuracy with the number of points per dimension for any problem.
    }
    \label{fig:all_res}
\end{figure}

%mention that there is no cross-correlations
We took the same approach to investigate the efficacy of GH and MC in all the test problems for $m=2$ and $3$. The results of the comparison are summarised in Figure \ref{fig:all_res}. We observed the same trends that with the increase in the number of points per dimension, we increase the accuracy. Even with a small number of points per dimension we are able to derive coefficients of over $0.85$ for all the problems. Interestingly, in some instances, e.g., WFG3 ($m=3$) and WFG4 ($m=2$), we clearly get better approximations from GH in comparison to MC.%\mnote{RA: Any idea why this is the case because the observation does not hold for the 2D and 3D counterparts of these two problems.}.

\subsection{Correlated multivariate Gaussian distribution}

The key issue with the analytical formula for EHVI is that it does not cater for correlated multivariate predictive distributions. However, both MC and GH, even though they are computationally relatively intensive, do not suffer from this issue. To investigate the efficacy of different methods, again, we take the same approach as before. We generate random distributions and compute the EHVI values with A, MC and GH, and then evaluate the rank correlations using Kendall's $\tau$ coefficient. Importantly, the most reliable method in this case is MC.

In this instance, the process to generate a random mean vector remains the same. However, for a valid covariance matrix, we must ensure that the randomly generated matrix remains positive definite. We, therefore, use Wishart distribution \cite{wishart1928generalised} to generate a positive definite matrix that is scaled by $\diag(u_1 - l_1, \dots, u_m - l_m)$. To demonstrate that the analytical version for uncorrelated distributions generates a poor approximation for the EHVI due to a correlated distribution, we use the diagonal of the covariance matrix and ignore the off-diagonal elements, and compute the EHVI. This allows us to quantitatively show that GH may be a better alternative to MC from an accuracy perspective.  

\begin{figure}[t!]
     \centering
     \begin{subfigure}[b]{0.32\textwidth}
         \centering
         \includegraphics[width=1\textwidth,trim={0mm 0mm 0mm 20mm}, clip=true]{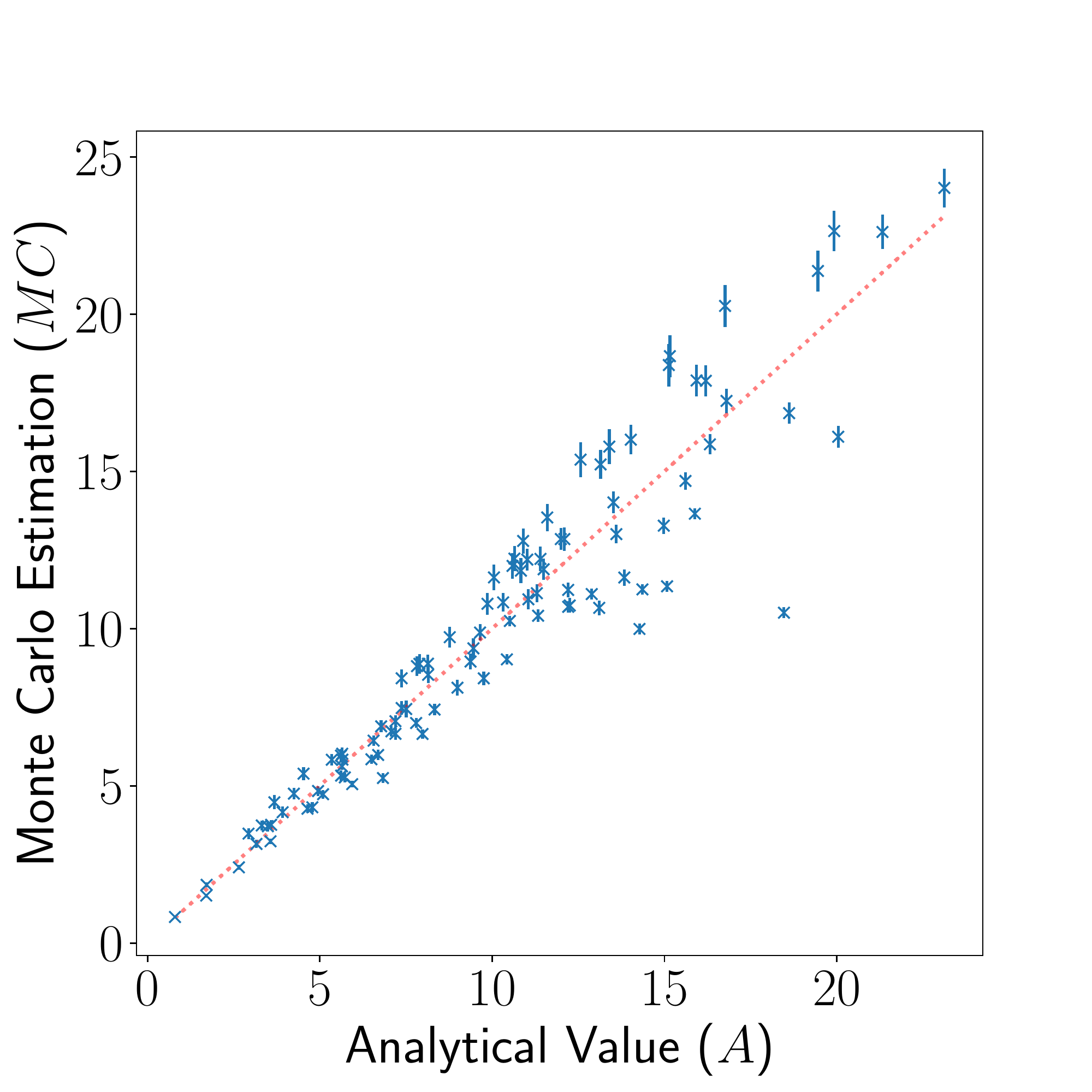}
         \caption{}
         \label{fig:mc_v_meas_corr}
     \end{subfigure}
     \hfill
     \begin{subfigure}[b]{0.32\textwidth}
         \centering
         \includegraphics[width=1\textwidth,trim={0mm 0mm 0mm 20mm}, clip=true]{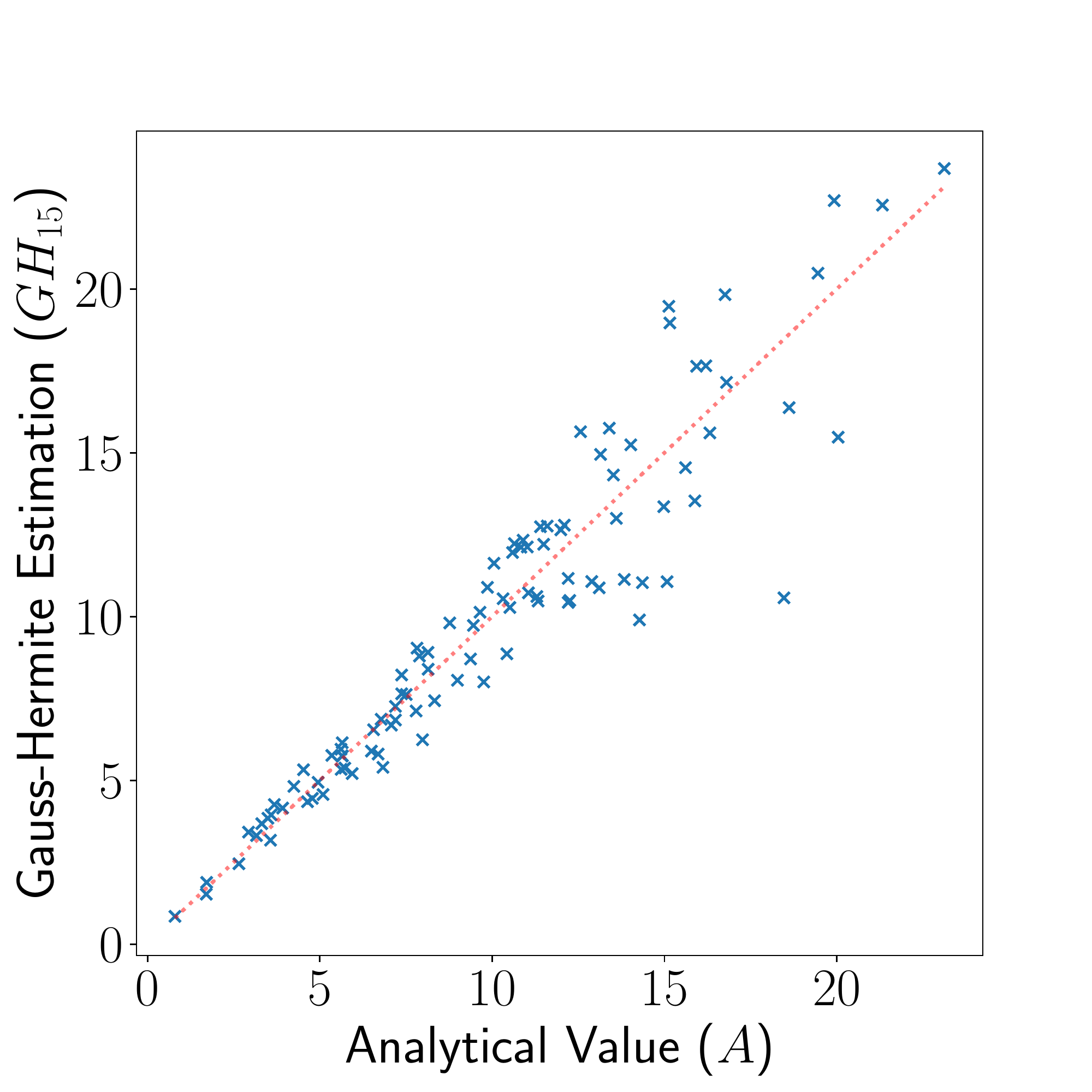}
         \caption{}
         \label{fig:gh_v_meas_corr}
     \end{subfigure}
     \hfill
     \begin{subfigure}[b]{0.32\textwidth}
         \centering
         \includegraphics[width=1\textwidth,trim={0mm 0mm 0mm 20mm}, clip=true]{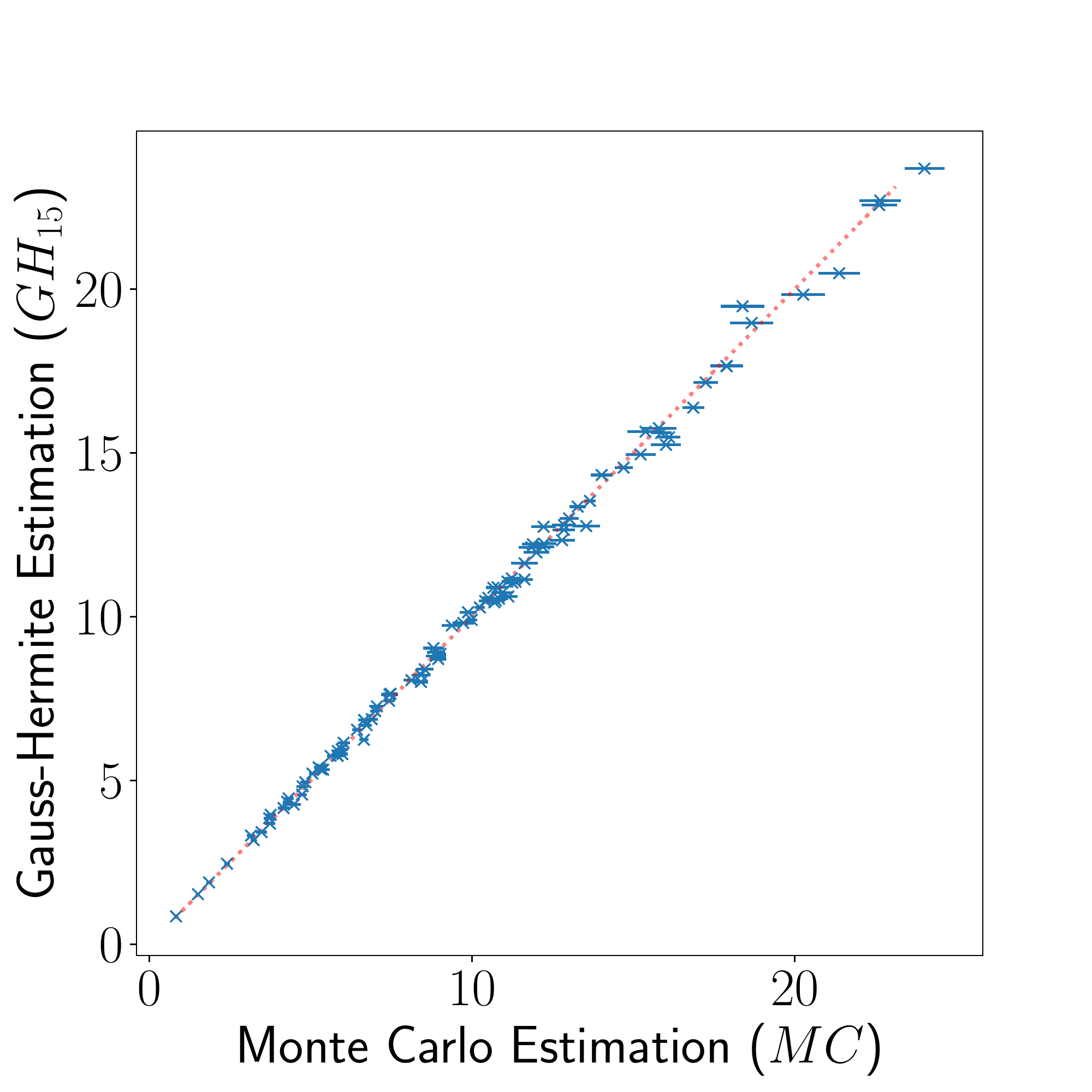}
         \caption{}
         \label{fig:gh_v_mc_corr}
     \end{subfigure}
     \hfill
     \begin{subfigure}[b]{1\textwidth}
         \centering
         \includegraphics[width=1\textwidth,trim={0mm 0mm 0mm 10mm}, clip=true]{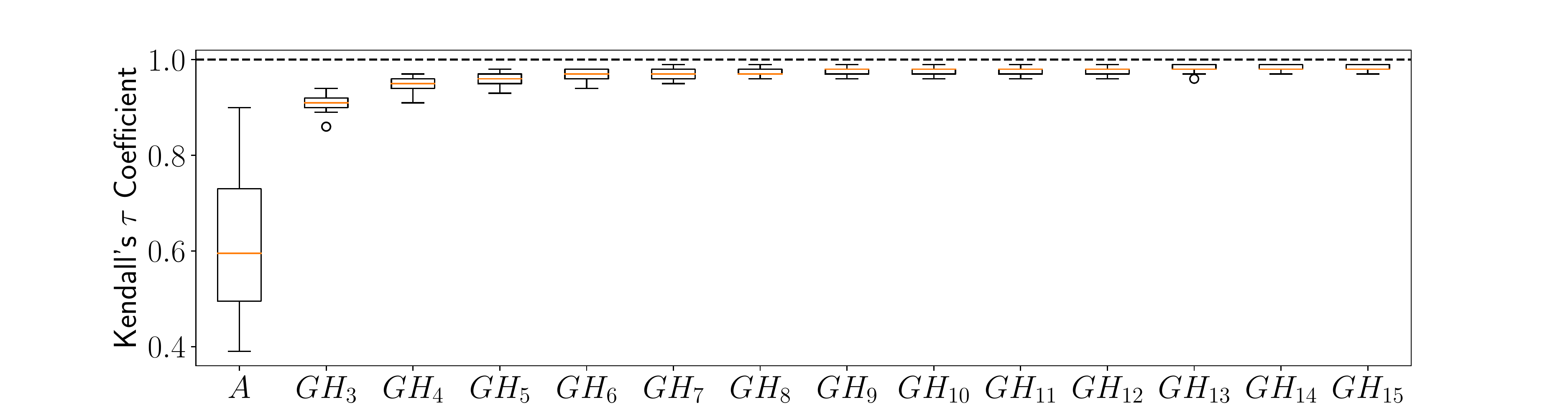}
         \caption{}
         \label{fig:all_res_corr}
     \end{subfigure}
        \caption{Illustration of the efficacy of GH for correlated multivariate Gaussian distributions as Figure \ref{fig:m_g_a} in \ref{fig:mc_v_meas_corr} -- \ref{fig:gh_v_mc_corr} for DTLZ2 ($m=2$). In \ref{fig:all_res_corr}, we show the summary of efficacies for different approximation methods when compared to MC across all DTLZ and WFG problems (for $m=2$ and $m=3$). Analytical approximations were generated using the diagonal of the covariance matrix. 
        % Illustration of the efficacy of GH for correlated multivariate Gaussian distributions (as Figure \ref{fig:m_g_a}). Analytical approximations were generated using the diagonal of the covariance matrix.
        %The analytical approach does not estimate the EHVI accurately, and as such the rank correlation with MC and GH are low, around $0.84$ (in \ref{fig:mc_v_meas_corr} and  \ref{fig:gh_v_meas_corr} respectively). However, both MC and GH agree significantly with each other with a rank correlation of $0.97$ (in \ref{fig:gh_v_mc_corr})
        \label{fig:m_g_a_corr}}
\end{figure}

In Figures \ref{fig:mc_v_meas_corr} -- \ref{fig:gh_v_mc_corr}, we show the comparison between different methods for computing EHVI for the DTLZ2 problem with $m=2$. Here, $A$ somewhat agrees with MC and $GH_{15}$ with a correlation coefficient of approximately $0.84$ in each case. However, MC and $GH_{15}$ are essentially producing the same ranking of solutions with a coefficient of just over $0.97$. Therefore, $GH_{15}$ with $180$ nodes is an excellent alternative to MC with $10,000$ samples. The results on DTLZ2 do not appear too bad for $A$. To ensure that this is the case for all test problems under scrutiny, we repeated the experiments, but this time generating $100$ random multivariate \textit{correlated} Gaussian distributions in each instance. Here, we assumed that MC is the most reliable measure, and computed the Kendall's $\tau$ coefficient with respect to MC. The correlation coefficient distributions for A and $GH_n$s are given in Figure \ref{fig:all_res_corr}. Clearly, there is a wide variance in the performance of A, with the minimum being $0.39$ for WFG7 ($m=3$) and maximum being $0.9$ for DTLZ1 ($m=2$). On the other hand, $GH_3$ produced the worst performance for GH across the board, but that was at $0.86$ for DTLZ ($m=2$), which shows a strong rank correlation. This shows that just considering the diagonal of the covariance matrix and computing the analytical EHVI is not a reliable approximation method under a \textit{correlated} multivariate predictive density. Instead, GH can produce a solid approximation with very few points.

\section{Conclusions}
\label{sect:concl}
EHVI is a popular acquisition function for expensive multi-objective optimisation. %due its reliance on the strictly Pareto compliant hypervolume metric.
Computing it analytically %for any number of objectives 
is possible for independent objectives (predictive densities). However, this can be prohibitively expensive for more than 3 objectives. Monte Carlo approximation can be used instead, but this is not cheap. We proposed an approach using GH quadrature as an alternative to approximating EHVI. Our experimental study showed that GH can be an accurate alternative to MC for both independent and correlated predictive densities with statistically significant rank correlations for a range of popular test problems. Future work can look at improving the computational efficiency of GH for high-dimensional problems, and validating GH within BO using EHVI as the acquisition function. 

%There are, of course, many ways in which this work can be extended. For instance, future work can look improving the computational efficiency of GH for high-dimensional problems using e.g. sparse grid quadrature~\cite{holtz2008sparse}. It is also important to validate GH within BO using EHVI as the acquisition function. 
% \input{Conclusions}

\section*{Acknowledgements}
This work is a part of the thematic research area Decision Analytics Utilizing Causal Models and Multiobjective
Optimization (DEMO, jyu.fi/demo) at the University of Jyvaskyla. Dr. Rahat was supported by the Engineering and Physical Research Council [grant number EP/W01226X/1]. 

%\section*{Acknowledgement}

\bibliographystyle{splncs04}
\bibliography{references}

\end{document}